\documentclass{bmvc2k}
\usepackage{bm}
\usepackage{amsfonts}
\usepackage{booktabs}

\title{Feature Fusion Vision Transformer for Fine-Grained Visual Categorization}

\addauthor{Jun Wang }{jun.wang.3@warwick.ac.uk}{1}
\addauthor{Xiaohan Yu }{xiaohan.yu@griffith.edu.au}{2}
\addauthor{Yongsheng Gao }{yongsheng.gao@griffith.edu.au}{2}

 \addinstitution{
 University of Warwick, UK
 }

\addinstitution{
 Griffith University, Australia\\
}
 \addinstitution{
  Collaborators, Inc.\\
  123 Park Avenue,\\
 }

\runninghead{}{}

\begin{document}

\maketitle

\begin{abstract}
The core for tackling the fine-grained visual categorization (FGVC) is to learn subtle yet discriminative features. Most previous works achieve this by explicitly selecting the discriminative parts or integrating the attention mechanism via CNN-based approaches. However, these methods enhance the computational complexity and make the model dominated by the regions containing the most of the objects. Recently, vision transformer (ViT) has achieved SOTA performance on general image recognition tasks. The self-attention mechanism aggregates and weights the information from all patches to the classification token, making it perfectly suitable for FGVC. Nonetheless, the classification token in the deep layer pays more attention to the global information, lacking the local and low-level features that are essential for FGVC. In this work, we propose a novel pure transformer-based framework 
Feature Fusion Vision Transformer (FFVT) where we aggregate the important tokens from each transformer layer to compensate the local, low-level and middle-level information. We design a novel token selection module called mutual attention weight selection (MAWS) to guide the network effectively and efficiently towards selecting discriminative tokens without introducing extra parameters. We verify the effectiveness of FFVT on four benchmarks where FFVT achieves the state-of-the-art performance. Code is available at \href{https://github.com/Markin-Wang/FFVT}{this link}.
\end{abstract}

\section{Introduction}
\label{sec:intro}
Fine-grained visual categorization (FGVC) aims to solve the problem of differentiating subordinate categories under the same basic-level category, e.g., birds, cars and plants. FGVC has wide real-world applications, such as autonomous driving and intelligent agriculture. Some FGVC tasks are exceedingly hard for human beings due to the small inter-class variance and large intra-class variance, e.g., recognizing 200 subordinate plant leaves and 200 subordinate birds. Therefore, FGVC is an important and highly challenging task.
\par

Owing to the decent designed networks and large-scale annotated datasets, FGVC has gained steady improvements in recent years. Current methods on FGVC can be roughly divided into localization-based methods and attention-based methods. The core for solving FGVC is to learn the discriminative features in images. Early localization-based methods \cite{huang2016part,berg2013poof,xie2013hierarchical} achieve this by directly annotating the discriminative parts in images. However, it is costly and time-consuming to build bounding box annotations, hindering the applicability of these methods on real-world applications. To alleviate this problem, recent localization-based methods normally integrate the region proposal network (RPN) to obtain the potential discriminative bounding boxes. These selected proposals are then fed into the backbone network to gain the local features. After that, most methods often adopt a rank loss \cite{chen2009ranking} on  the classification outputs for all local features. However, \cite{he2021transfg} argues that RPN-based methods ignore the relationships among selected regions. Another problem is that this mechanism drives the RPN to propose large bounding boxes as they are more likely to contain the foreground objects. Confusion occurs when these bounding boxes are inaccurate and cover the background rather than objects. Besides, some discriminative regions, e.g., leaf vein in plant leaves, cannot be simply annotated by a rectangular \cite{9506424}.
\par
Attention-based \cite{zhao2017diversified, xiao2015application, zheng2021rethinking} methods automatically detect the discriminative regions in images via self-attention mechanism. These methods release the reliance on manually annotation for discriminative regions and have gained encouraging results. Recently, vision transformer has demonstrated potential performance on general image classification \cite{dosovitskiy2020image}, image retrieval \cite{el2021training} and semantic segmentation \cite{zheng2021rethinking}. This great success shows that the innate attention mechanism of a pure transformer architecture can automatically search the important parts in images that contribute to image recognition. However, few study investigate the performance of vision transformer in FGVC. As the first work to study the vision transformer on FGVC, \cite{he2021transfg} proposed to replace the inputs of the final transformer layer with some important tokens and gained improved results. Nonetheless, the final class token may concern more on global information and pay less attention to local and low-level features, defecting the performance of vision transformer on FGVC since local information plays an important role in FGVC. Besides, previous works focus on FGVC benchmarks containing more than ten thousands of annotated images, and no study explores the capability of vision transformer on small-scale and ultra-fine-grained visual categorization (ultra-FGVC) settings.
\par  
In this paper, we propose a novel feature fusion vision transformer (FFVT) for FGVC. FFVT aggregates the local information from low-level, middle-level and high-level tokens to facilitate the classification. We present a novel important token selection approach called Mutual Attention Weight Selection (MAWS) to select the representative tokens on each layer that are added as the inputs of the last transformer layer. In addition, we explore the performance of our method on four FGVC datasets to comprehensively verify the capability of our proposed FFVT on FGVC. In conclusion, our work has four main contributions. 
\par
1. To our best knowledge, we are the first study to explore the performance of vision transformer on both small-scale and ultra-FGVC settings. The two small-scale datasets in this paper are highly challenging due to the ultra-fine-grained inter-category variances and few training data available. Some examples are visualized in Figure 1.
\par
2. We propose FFVT, a novel vision transformer framework for fine-grained visual categorization tasks that can automatically detect the distinguished regions and take advantage of different level of global and local information in images.
\par
3. We present a novel important token selection approach called Mutual Attention Weight Selection (MAWS). MAWS can effectively select the informative tokens that are having  high similarity to class token both in the contexts of the class token and the token itself without introducing extra parameters. 
\par
4. We verify the effectiveness of our method on four fine-grained benchmarks. Experimental results demonstrate that FFVT achieves state-of-the-art performance on them, offering an alternative to current CNN-based approaches. Ablation studies show that our proposed method boost the performance of the backbone model by 5.42\%, 4.67\% and 0.80\% on  CottonCultivar80, SoyCultivarLocal and CUB datasets, respectively. 

\begin{figure*}[ht]   
    \centering 
    \includegraphics[width=0.7\textwidth]{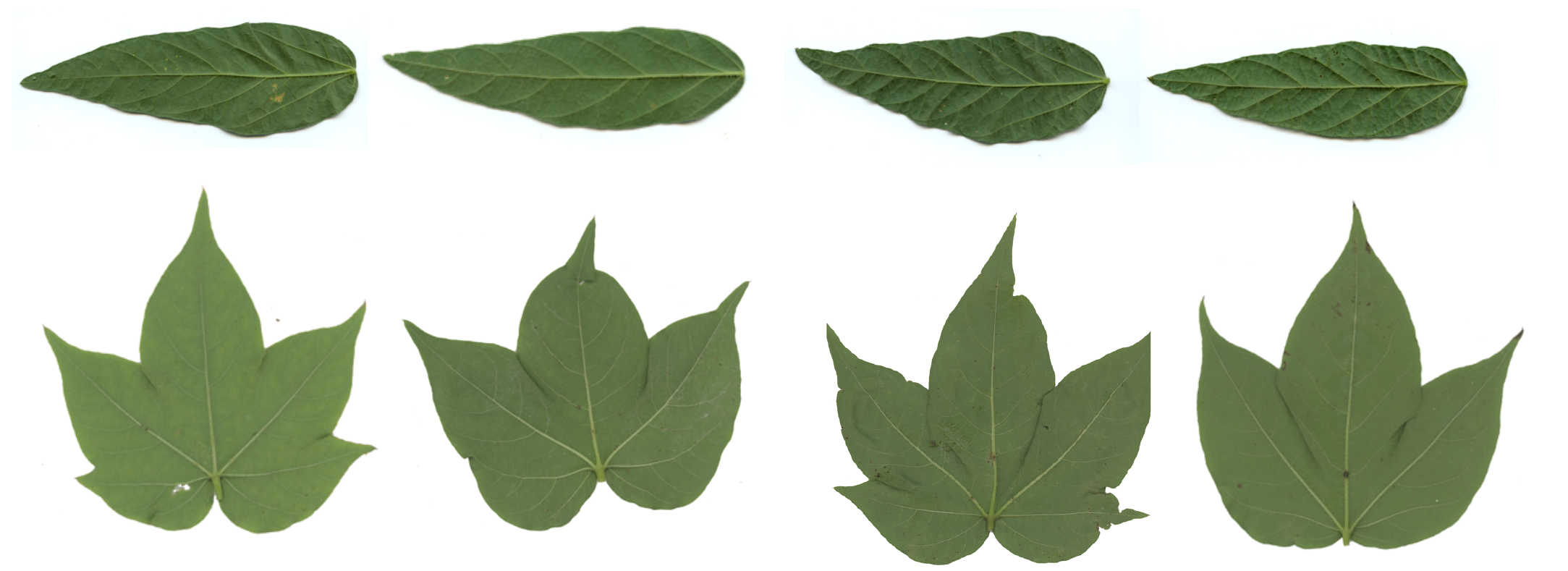} 
    \caption{Examples of images in SoyCultivarLocal and Cotton datasets. Images in the first row come from four species of Soy.Loc, while examples in the second row are selected from four categorizes of Cotton. }    
    \label{Leaves_Examples}  
\end{figure*}

\section{Related Works}
\subsection{Fine-Grained Visual Categorization}
Methods on FGVC can be coarsely divided into two groups: localization-based methods and attention-based methods. Similar to object detection task, localization-based methods often detect the foreground objects and perform classification based on them. Early works \cite{huang2016part,berg2013poof,xie2013hierarchical} achieve this by taking advantage of part annotation to supervise the learning of the detection branch. However, bounding box annotation requires large manual labor, hampering their real-world applications. 
\par
To alleviate above problem, recent localization-based methods introduce the weakly supervised object detection (WSOD) technique to predict the potential discriminative regions with only image-level label. Ge et al. \cite{ge2019weakly} used WSOD and instance segmentation techniques to obtain the rough object instances, and then selected the important instances to perform classification. He et al. \cite{he2017weakly} presented two spatial constraints to select the discriminative parts obtained by the detection branch. Wang et al. \cite{wang2019weakly} utilized correlations between regions to select distinguished parts. However, these methods require a well designed WSOD branch to propose potential discriminative regions. Moreover, the selected parts sent to the classification head often cover the whole object instead of the truly discriminative parts.
\par
Alternatively, attention-based methods automatically localize the discriminative regions via self-attention mechanism without extra annotations. Zhao et al. \cite{zhao2017diversified} proposed a diversified visual attention network which uses the diversity of the attention to collect dicriminative information. Xiao et al. \cite{xiao2015application} presented a two-level attention mechanism to steadily filter out the trivial parts. Similar to \cite{xiao2015application}, Zheng et al. \cite{zheng2021rethinking} proposed a progressive-attention to progressively detect discriminative parts at multiple scales. However, these methods often suffer from huge computational cost. 

\subsection{Transformer}
Transformer has achieved huge success in natural language processing \cite{devlin2018bert,tsai2019multimodal,vaswani2017attention}. Motivated by this, researchers try to exploit the transformers in computer vision. Recent work ViT \cite{dosovitskiy2020image} achieves the state-of-the-art performance on image classification by employing a pure transformer architecture on a number of fix-sized image patches. Later, researchers explore the performance of the pure transformer in other computer vision tasks. Zheng \cite{zheng2021rethinking} et al. developed a pure transformer SETR on semantic segmentation task. Alaaeldin et al. \cite{el2021training} exploited a transformer to generate the image descriptor for image retrieval task. Nonetheless, few studies explore the vision transformer on FGVC. \par 
The most similar to our work is TransFG \cite{he2021transfg} which is the first study to extend the ViT into FGVC, while there are two notable differences between FFVT and TransFG. First, TransFG selects the discriminative tokens and directly send them to the last transformer layer (no feature fusion), while FFVT aims to aggregate the local and different level information from each layer to enrich the feature representation capability via feature fusion. Second, our proposed token selection strategy is totally different from that of TransFG which requires the attention information from all transformer layer to generate the selected token indexes via matrix multiplication. In contrast, our proposed MAWS utilize attention information from only one transformer layer to produce the corresponding indexes. Hence, MAWS is simple and efficient. Our work is also in accordance with the spirit of recent research \cite{YUICCV21,YU2021108067,wang2021two, ZHAO2021107938,yu2020patchy,wang2021boosted,zhao2022learning,yu2019multiscale,yu2015leaf,yu2016multiscale,9506424, wang2021ear,zhang2009local,gao2002face}, which focuses on localizing subtle yet vital regions.

\section{Methods}
To better comprehend our method, we first briefly review the knowledge of vision transformer ViT in Section 3.1. Our proposed methods are then elaborately described in the following subsections.
\par

\subsection{ViT For Image Recognition}
ViT follows the similar architecture of transformer in natural language processing with minor modification. Transformer in natural language processing takes a sequence of tokens as inputs. Similarly, given an image with resolution $ H*W$,  vision transformer first processes the image into $N=\lfloor \frac{H}{P} \rfloor * \lfloor \frac{W}{P} \rfloor$ fix-sized patches $x_p$. where $P$ is the size for each patch. 
\par
The patches $x_p$ are then linearly projected into a D-dimensional latent embedding space. To introduce the positional differences, a learnable vector called position embedding with the same size of patch embedding is directly added to patch embedding. Similar to the class token in BERT \cite{devlin2018bert}, an extra class token is added to interact with all patch embeddings and undertakes the classification task. The procedure is shown in Eq (1):
\begin{equation}
 \bm{z}_0=[\bm{x}_{class};\ \bm{x}^1_p\bm{E};\ \bm{x}^2_p\bm{E};\ ...;\ \bm{x}^N_p\bm{E}]+\bm{E}_{pos}
\end{equation}
\noindent
Where $\bm{E}\in \mathbb{R}^{(P^2\cdot C) \times D}$ is the patch embedding projection and $C$ is the number of the image channels. $\bm{E}_{pos}$ denotes the position embedding.
\par
After that, these patch embeddings are fed into the transformer encoder containing several multi-head self-attention (MSA) and multi-layer perceptron (MLP) blocks. Note that all layers retain the same latent vector size $D$. The outputs of the $l$-th layer are calculated by Eqs (2) to (3):
\begin{equation}
 \bm{z}^{'}_{l}=MSA(LN(\bm{z}_{l-1}))+\bm{z}_{l-1}
\end{equation}
\begin{equation}
 \bm{z}_{l}=MLP(LN(\bm{z}^{'}_{l}))+\bm{z}^{'}_{l}.
\end{equation}
\noindent
Where $LN(\cdot)$ is the layer normalization operation and $z_l$ denotes the encoded image representation. Eventually, a classification head implemented by a MLP block is applied to the class token $\bm{z}_l^0$ to obtain the predicted category.
\par

\subsection{FFVT Architecture}
\cite{he2021transfg} suggests that the ViT cannot capture enough local information required for FGVC. To cope with this problem, we propose to fuse the low-level features and middle-level features to enrich the local information. We present a novel token selection approach called mutual attention weight selection (MAWS) to determine the tokens to be aggregated in the deep layer. This section introduces the details of our proposed FFVT. The overall architecture of FFVT is illustrated in Fig 2.

\begin{figure*}[ht]   
    \centering 
    \includegraphics[width=0.8\textwidth]{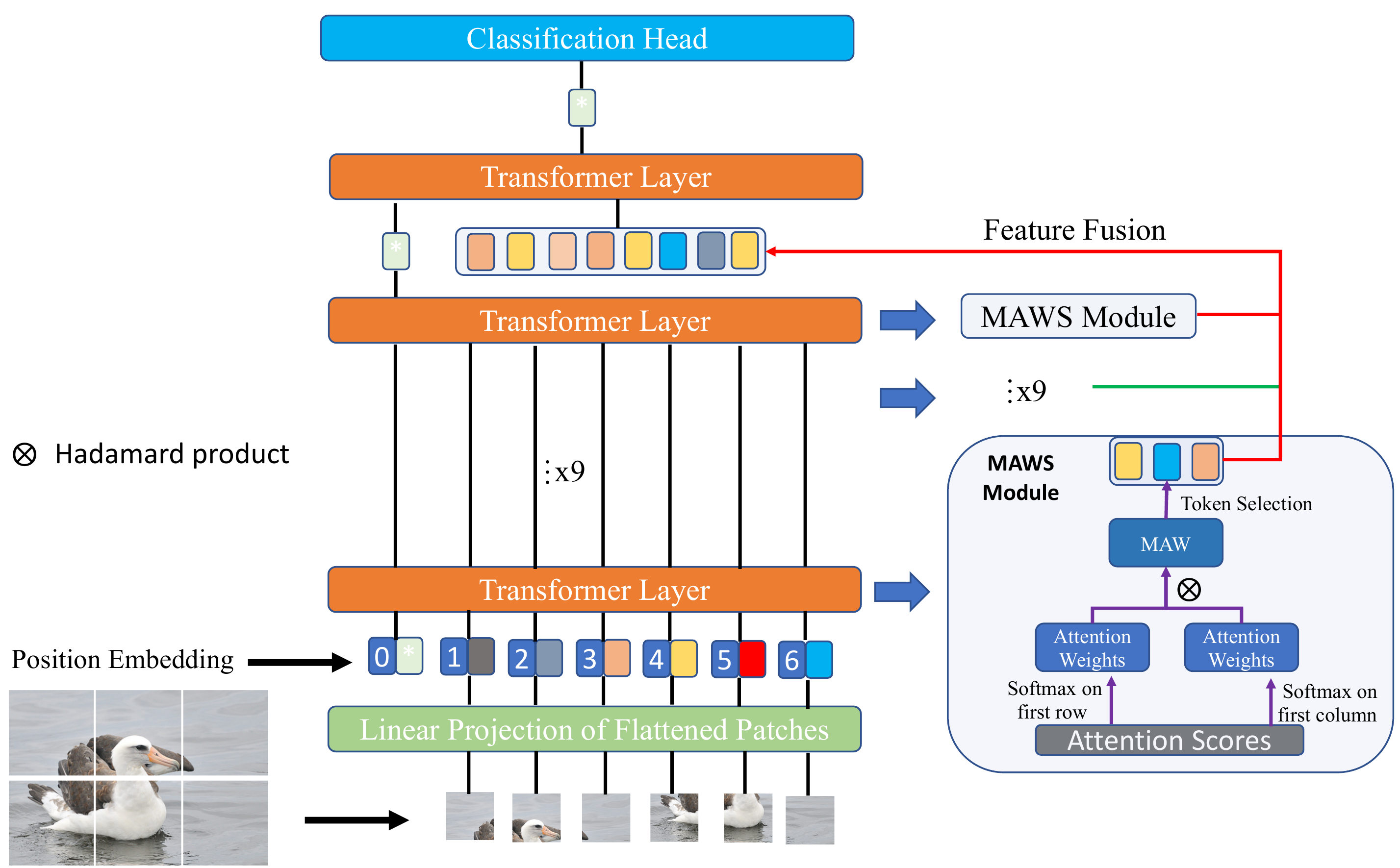} 
    \caption{The overall architecture of the proposed FFVT. Images are split into a sequence of fix-sized patches which are then linearly projected into the embedding space. Combined with the position embedding, the patch embeddings are fed into the Transformer Encoder to learn the patch features. Feature fusion is exploited before the last transformer layer to aggregate the important local, low-level and middle level information from previous layers. This is implemented by replacing the inputs (exclude classification token) of the last transformer layer with the tokens selected by the MAWS Module. }    
    \label{fig_overall_architecture}  
\end{figure*}

\subsubsection{Feature Fusion Module}
The key challenge of the FGVC is to detect the discriminative regions that significantly contribute to figuring out the subtle differences among subordinate categories. Previous works often achieve this by manually annotating the discriminative regions or integrating the RPN module. However, these methods suffer from some problems discussed in Section 1\&2, limiting their performance on real-world applications.
\par
The MSA mechanism in vision transform can perfectly meet the above requirement, whereas MSA in deep layer is likely to pay more attention to the global information. Therefore, we propose a feature fusion module to compensate local information. As shown in figure 2, given the important tokens (hidden features) from each layer selected by MAWS module, we replace the inputs (except for the class token) of the last transformer layer with our selected tokens. In this way, the class token in the last transformer layer fully interacts with the low-level, middle level and high-level features from the previous layers, enriching the local information and feature representation capability. 
\par
Specifically, we denote the tokens selected by MAWS module in $l$-layer as:
\begin{equation}
 \bm{z}_l^{local}=[z_l^1,\ z_l^2,\ ...,\ z_l^K]
\end{equation}
Where $K$ is the number of selected features. The fused features along with the classification token fed into the last transformer layer $L$ are:
\begin{equation}
 \bm{z}_{ff}=[\bm{z}_{L-1}^0;\  \bm{z}_1^{local};\  \bm{z}_2^{local};  ...;\  \bm{z}_{L-1}^{local}]
\end{equation}
\par
Eventually, following the ViT, the classification token of the final transformer layer is sent to the classification head to perform categorization. The problem turns to how to select the important and discriminative tokens. To that end, we propose an effective and efficient token selection approach described in the next section.

\subsubsection{Mutual Attention Weight Selection Module}
Since an image is split into many patches, token selection turns to be an important problem. Noise is added when the background patches are frequently selected, while discriminative patches can boost the model performance. Hence, we propose a token selection approach which directly utilizes the attention scores generated by multi-head self-attention module.
\par
To be specific, an attention score matrix for one attention head $A \in \mathbb{R}^{(N+1)\times (N+1)}$ is denoted as:
\begin{equation}
 \bm{A}=[\bm{a}^0;\  \bm{a}_1;\  \bm{a}_2;\ ...;\ \bm{a}_{i};\ ...;\ \bm{a}_{N}]
\end{equation}
\begin{equation}
 \bm{a}_i=[a_{i,0},\ a_{i,1},\ a_{i,2}, \ ...,\ a_{i,j},\ ...,\  a_{i,N} ]
\end{equation}
Where $a_{i,j}$ is the attention score between token $i$ and $j$ in the context of token $i$, i.e., dot-product between the query of token $i$ and the key of token $j$.
\par
One simple strategy is to pick the tokens having the higher attention scores with the classification token as the classification token contains rich information for categorization. We can do this by sorting the $\bm{a}_0$ and picking the $K$ tokens with the bigger value. We denote this strategy as single attention weight selection (SAWS). However, SAWS may introduce noisy information since the selected tokens could aggregate much information from noisy patches. Taking a three-patch attention score matrix $\gamma$ shown below as an example:
$$
\gamma=
\left[
\begin{matrix}
   \bm{a}^0  \\
   \bm{a}^1 \\
   \bm{a}^2  \\
   \bm{a}^3 
   \end{matrix}
   \right]
=
\left[
\begin{matrix}
   1 & 2 & 3 & 4 \\
   1 & 2 & 3 & 4 \\
   1 & 2 & 3 & 4 \\
   1 & 4 & 1 & 1
   \end{matrix}
   \right]
$$
\par
Token three is selected as it has the biggest value in the attention score vector for classification token. However, token three aggregates much information from token one (the maximum attention score in $\bm{a}_3$) thus may introduce noises assuming token one is a noisy token. To cope with this problem, we develop a mutual attention weight selection module which requires the selected tokens to be similar to the classification token both in the contexts of the classification token and the tokens themselves.
\par
In particular, we denote the first column in the attention score matrix as $\bm{b}_0$. Note that $\bm{b}_0$ is the attention score vector between classification token and other tokens in the context of \textbf{other tokens} compared with $\bm{a}_0$ in the context of \textbf{classification token}. The mutual attention weight $\bm{ma}_i$ between the classification token and token $i$ is then calculated by Eqs (8) to (9):
\begin{equation}
 \bm{ma}_i=a^{'}_{0,i}*b^{'}_{i,0}
\end{equation}
\begin{equation}
 a^{'}_{0,i}=\frac{e^{a_{0,i}}}{\sum_{j=0}^{N}e^{a_{0,j}}}, \   
 b^{'}_{i,0}=\frac{e^{b_{i,0}}}{\sum_{j=0}^{N}e^{b_{j,0}}}
\end{equation}
\par
For multi-head self-attention, we first average the attention scores of all heads. After obtaining the mutual attention weight, the indexes of important tokens are collected according to the mutual attention values. Our approach does not introduce extra learning parameters. It is simple and efficient compared with the matrix multiplication in \cite{he2021transfg}.

\section{Experiments}

\subsection{Datasets}
We explore the effectiveness of FFVT on two widely used FGVC dataset and two small-scale ultra-fine-grained datasets, i.e., CUB-200-2011 \cite{wah2011caltech}, Stanford Dogs \cite{khosla2011novel}, SoyCultivarLocal \cite{yu2020patchy} and CottonCultivar80 \cite{yu2020patchy}. The SoyCultivarLocal  and CottonCultivar80  are two highly challenging datasets as they further reduce the granularity of categorization, e.g. from species to cultivar, and with few training data available. The statistics of four datasets are shown in Table 1.

\begin{table}
\begin{center}
\begin{tabular}{|c|c|c|c|}
\hline
Datasets & Category & Training & Testing \\
\hline
CUB-birds & 200 & 5994 & 5794 \\
\hline
Stanford Dogs & 196 & 8144 & 8041 \\
\hline
Soy.Loc & 200 & 600 & 600 \\
\hline
Cotton & 80 & 240 & 240 \\
\hline
\end{tabular}
\end{center}
\caption{Statistics of CUB-200-2011, SoyCultivarLocal (Soy.Loc), CottonCultivar80 (Cotton) and Stanford Dogs datasets.}
\end{table}

\subsection{Implementation Details}
The same as the most current transformer-based approaches, the backbone network (ViT) of FFVT is pretrained on the ImageNet21K dataset. Following the same data augmentation methods on most existing works, input images are first resized to $500 \times 500$ for Soy.Loc and Cotton datasets, and $600 \times 600$ for CUB and Stanford Dogs. We then crop the image into $384 \times 384$ for Soy.Loc and Cotton, and $448 \times 448$ for CUB and Stanford Dogs (Random cropping in training and center cropping in testing). Random horizontal flipping is adopted and an extra color augmentation is applied for CUB. $K$ in Eq (4) is set to 12 for CUB, Soy.Loc and Cotton, and 24 for Stanford Dogs. 
\par
We select the SGD optimizer to optimize the network with a momentum of 0.9. The initial learning rate is 0.02 with the cosine annealing scheduler for FFVT on CUB, Soy.Loc Cotton datasets, and 0.003 on the Stanford Dogs dataset. The batch size is set to 8 for all datasets except for the Stanford Dogs with a batch size of 4. For fair comparisons, we reimplement the experiments of ViT and TransFG on the Stanford Dogs benchmark with their default settings and the same batch size as FFVT. Experiments are conducted on four Nvidia 2080Ti GPUs using PyTorch deep learning framework.

\subsection{Comparison with the State-Of-The-Art}
Here, we demonstrate the experimental results on four datasets and compare our method with a number of state-of-the-art works. As shown in Table 2, FFVT obtains the second best-performed method on CUB with an accuracy of 91.6\%, beating other methods by a large margin except for the most recent state-of-the-art fine-grained method TransFG (-0.1\%). Note that FFVT achieves a comparable accuracy against TransFG with much less computation cost and GPU memory consumption since the overlapping strategy of TransFG significantly increases the number of the input patches from 784 to 1296. Besides, limited by our computation resources, the batch size of TransFG on the experiment of CUB dataset is two times larger than FFVT. This may also account for the relative performance differences. FFVT outperforms all the listed approaches on Stanford Dogs with an accuracy of 91.5\%, strongly exceeding the second best-performed TransFG by 0.9\%.

\begin{table}[!h]
\caption{Comparison of different methods on CUB-200-2011 datasets. The best accuracy is highlighted in bold and the second best accuracy is underlined.}
\centering
\begin{tabular}{c|c|c}
\toprule  
Method& Backbone& Accuracy \\
\hline
ResNet50 \cite{he2017weakly}&  ResNet50&  84.5\\
GP-256 \cite{wei2018grassmann}& VGG16 & 85.8\\
MaxEnt \cite{dubey2018maximum}& DenseNet161 & 86.6\\
DFL-CNN \cite{wang2018learning}&  ResNet50 & 87.4\\
NTS-Net \cite{yang2018learning}&  ResNet50 & 87.5\\
Cross-X \cite{luo2019cross}&  ResNet50 & 87.7\\
DCL \cite{chen2019destruction}&  ResNet50 & 87.8\\
CIN \cite{gao2020channel}&  ResNet101 & 88.1\\
DBTNet \cite{zheng2019learning}&  ResNet101 & 88.1\\
ACNet \cite{ji2020attention}&  ResNet50 & 88.1\\
S3N \cite{ding2019selective}&  ResNet50 & 88.5\\
FDL \cite{liu2020filtration}&  DenseNet161 & 89.1\\
PMG \cite{du2020fine}&  ResNet50 & 89.6\\
API-Net \cite{zhuang2020learning}&  DenseNet161 & 90.0\\
StackedLSTM \cite{ge2019weakly}&  GoogleNet & 90.4\\
\hline
ViT \cite{dosovitskiy2020image}&  ViT-B\_16&  90.8\\

TransFG \cite{he2021transfg}&  ViT-B\_16 &  \textbf{91.7}\\
FFVT&  ViT-B\_16 &  \underline{91.6}\\
\bottomrule 
\end{tabular}
\end{table}

\begin{table}[!h]
\caption{Comparison of different methods on Stanford Dogs (Dogs) dataset. The best accuracy is highlighted in bold and the second best accuracy is underlined. Values in parentheses are reported results in their papers.}
\centering
\begin{tabular}{c|c|c}
\toprule  
Method& Backbone& Dogs\\
\hline
MaxEnt \cite{dubey2018maximum}& DenseNet161 & 83.6\\
FDL \cite{liu2020filtration}&  DenseNet161 & 84.9\\
RA-CNN \cite{fu2017look}&  VGG19 & 87.3\\
DB \cite{sun2020fine}&  ResNet50 & 87.7\\
SEF \cite{luo2020learning}&  ResNet50 & 88.8\\
Cross-X \cite{luo2019cross}&  ResNet50 & 88.9\\
API-Net \cite{zhuang2020learning}&  DenseNet161 & 90.3\\
\hline
ViT \cite{dosovitskiy2020image}&  ViT-B\_16&  90.2\\

TransFG \cite{he2021transfg}&  ViT-B\_16 &  \underline{90.6} (92.3)\\
FFVT&  ViT-B\_16 &  \textbf{91.5}\\
\bottomrule 
\end{tabular}
\end{table}

SoyCultivarLocal and CottonCultivar80 are two extremely challenging ultra-fine-grained datasets. The difficulty lies in two folds, i.e., super-subtle inter-class differences and few training images (three for each category). Some examples are visualized in figure 1. Therefore, locating the discriminative regions plays an essential role in accurate classification. The results of experiments on SoyCultivarLocal and CottonCultivar80 are shown in Table 3. FFVT obtains the highest accuracy of 57.92\% on CottonCultivar80, outperforming the second best-performed method by a large margin (+4.17\%). Similarly, our proposed FFVT beats all methods with an accuracy of 44.17\% on SoyCultivarLocal.
\par

\begin{table}[!h]
\caption{Comparison of different methods on SoyCultivarLocal (Soy.Loc) and CottonCultivar80 (Cotton) datasets. The best accuracy is highlighted in bold and the second best accuracy is underlined.}
\centering
\begin{tabular}{c|c|c|c}
\toprule  
Method& Backbone& Cotton & Soy.Loc \\
\hline
AlexNet \cite{krizhevsky2012imagenet}&  AlexNet&  22.92 & 19.50\\
VGG16 \cite{simonyan2014very}& VGG16& 50.83 & 39.33 \\
ResNet50 \cite{he2017weakly}& ResNet50 & 52.50 & 38.83\\
InceptionV3 \cite{szegedy2016rethinking}& GoogleNet & 37.50 & 23.00\\
MobileNetV2 \cite{sandler2018mobilenetv2}&  MobileNet & 49.58 & 34.67\\
Improved B-CNN \cite{lin2017improved}&  VGG16 & 45.00 & 33.33\\
NTS-Net \cite{yang2018learning}&  ResNet50 & 51.67 & \underline{42.67}\\
fast-MPN-COV \cite{li2018towards}&  ResNet50 & 50.00 & 38.17\\

\hline
ViT \cite{dosovitskiy2020image}&  ViT-B\_16&  51.25 & 39.33\\
DeiT-B \cite{touvron2020training} & ViT-B\_16 & \underline{53.75} & 38.67 \\
TransFG \cite{he2021transfg}&  ViT-B\_16 &  45.84 & 38.67\\
FFVT&  ViT-B\_16 &  \textbf{57.92} & \textbf{44.17}\\
\bottomrule 
\end{tabular}
\end{table}

\subsection{Ablation Studies}
We perform the ablation studies on CottonCultivar80, SoyCultivarLocal and CUB to further validate the effectiveness of our proposed methods. SAWS is the single attention weight selection strategy designed in Section 3.2.2. As shown in Table 5, even the simple SAWS strategy can remarkably boost the performance by 4.58\%, 3.50\% and 0.64\% on CottonCultivar80, SoyCultivarLocal and CUB, respectively. The results confirm the necessity of aggregating the local and different level information for vision transformer on FGVC. A bigger improvement can be seen when applying the MAWS strategy (+6.67\%, 4.84\% and 0.80\% on CottonCultivar80, SoyCultivarLocal and CUB, respectively), showing that MAWS better exploits the attention information. MAWS explicitly selects the most useful tokens thus forces the model to learn from these informative parts.

\begin{table}[!h]
\caption{Ablation studies on CottonCultivar80 (Cotton), SoyCultivarLocal (Soy.Loc), and CUB datasets. The best accuracy is highlighted in bold.}
\centering
\begin{tabular}{c|c|c|c}
\toprule  
Method& Cotton& Soy.Loc& CUB \\
\hline
ViT \cite{dosovitskiy2020image}&51.25&  39.33&  90.85\\
\hline
ViT+Feature Fusion+SAWS &55.83& 42.83&  91.49 \\
\hline
FFVT(ViT+Feature Fusion+MAWS)&\textbf{57.92} & \textbf{44.17} & \textbf{91.65} \\
\bottomrule 
\end{tabular}
\end{table}

We then investigate the influence of the hyper-parameter $K$. Table 6 summarizes the results of FFVT on the SoyCultivarLocal dataset with the value of $K$ ranging from 10 to 14. FFVT achieves the best performance when there are 12 tokens selected for each layer. One possible reason is that the tokens focused by each attention head are selected by the proposed MAWS module and contribute positively to the classification since this value (12) is in accordance with the number of the attention heads. As $K$ increases from 10 to 12, the accuracy steadily enhances from 43.17\% to 44.17\%. A different pattern can be seen when $K$ continues increasing to 14, where the accuracy slightly reduces to 42.5\%. The performance drop may due to that large $K$ introduces the noisy tokens while small $K$ value lead to insufficient discriminative information for classification. Note that results of all $K$ settings show a significant improvements over that backbone ViT (39.33\%), indicating that FFVT is not very sensitive to the value of $K$.

\begin{table}[!h]
\caption{Ablation studies of the hyper-parameter $K$ on SoyCultivarLocal benchmark. The best accuracy is highlighted in bold.}
\centering
\begin{tabular}{cccccc}
\toprule  
$K$ & 10& 11& 12& 13& 14 \\
\hlineß
Accuracy(\%) &43.17&  43.83& \textbf{44.17}& 43.00& 42.50\\

\bottomrule 
\end{tabular}
\end{table}

\section{Conclusion}
This paper proposes a novel fine-grained visual categorization architecture FFVT and achieves state-of-the-art performance on four benchmarks. To facilitate the performance of vision transformer in FGVC, we propose a feature fusion approach to enrich the local, low-level and middle-level information for the classification token. To select the discriminative tokens that to be aggregated, we develop a novel token selection module MAWS which explicitly takes advantage of the attention scores produced by self-attention mechanism. Experimental results show that FFVT significantly improve the classification accuracy of standard ViT on different fine-grained settings, i.e., normal-scale, small-scale and ultra-fine-grained settings. We observe that FFVT is very effective on the challenging datasets, confirming its capability of capturing subtle differences and discriminative information. 
\par
Based on our encouraging results, we believe that the pure-transformer model has the huge potential on different FGVC settings, even in the small-scale datasets without the induction bias like convolutional neural networks.

\bibliography{egbib}
\end{document}